\documentclass[sigconf]{acmart}

\usepackage{svg}

\usepackage{balance}
\usepackage{subcaption}
\usepackage{graphicx}
\usepackage{multirow}
\usepackage{hyperref}
\usepackage{colortbl}

\usepackage[]{algpseudocode}
\usepackage{algorithm}
\algnewcommand\algorithmicforeach{\textbf{for each}}
\algdef{S}[FOR]{ForEach}[1]{\algorithmicforeach\ #1\ \algorithmicdo}

\copyrightyear{2022}
\acmYear{2022}
\setcopyright{acmcopyright}\acmConference[GECCO '22]{Genetic and Evolutionary
Computation Conference}{July 9--13, 2022}{Boston, MA, USA}
\acmBooktitle{Genetic and Evolutionary Computation Conference (GECCO '22), July
9--13, 2022, Boston, MA, USA}
\acmPrice{15.00}
\acmDOI{10.1145/3512290.3528719}
\acmISBN{978-1-4503-9237-2/22/07}

\begin{document}

% \title{Decision Space Analysis of Automatically Designed Multiobjective Algorithms}
\title{Component-wise Analysis of Automatically Designed Multiobjective Algorithms on Constrained Problems}
\author{Yuri Lavinas}
\email{lavinas.yuri.xp@alumni.tsukuba.ac.jp}
\orcid{0000-0003-2712-5340}
\affiliation{%
  \institution{University of Tsukuba}
  \country{Japan}
}

\author{Marcelo Ladeira}
\email{mladeira@unb.br}
\affiliation{%
  \institution{University of Brasilia}
  \country{Brazil}
}

\author{Gabriela Ochoa}
\email{gabriela.ochoa@cs.stir.ac.uk}
\affiliation{%
  \institution{University of Stirling}
  \country{UK}
}

\author{Claus Aranha}
\email{caranha@cs.tsukuba.ac.jp}
\affiliation{%
  \institution{University of Tsukuba}
  \country{Japan}
}

\renewcommand{\shortauthors}{Lavinas, et al.}

\begin{abstract}
   The performance of multiobjective algorithms varies across problems, making it hard to develop new algorithms or apply existing ones to new problems. To simplify the development and application of new multiobjective algorithms, there has been an increasing interest in their automatic design from component parts. These automatically designed metaheuristics can outperform their human-developed counterparts. However, it is still uncertain what are the most influential components leading to their performance improvement. This study introduces a new methodology to investigate the effects of the final configuration of an automatically designed algorithm. We apply this methodology to a well-performing Multiobjective Evolutionary Algorithm Based on Decomposition (MOEA/D) designed by the irace package on nine constrained problems. We then contrast the impact of the algorithm components in terms of their Search Trajectory Networks (STNs), the diversity of the population, and the hypervolume. Our results indicate that the most influential components were the restart and update strategies, with higher increments in performance and more distinct metric values. Also, their relative influence depends on the problem difficulty: not using the restart strategy was more influential in problems where MOEA/D performs better; while the update strategy was more influential in problems where MOEA/D performs the worst.
\end{abstract}





\maketitle

\section{Introduction}
\label{section:intro}

Multiobjective Optimisation Problems (MOPs) are problems with two or more conflicting objective functions that are optimised simultaneously. 
The evolutionary computation community has proposed several multiobjective evolutionary algorithms (MOEAs), which can modify their behaviour when solving a MOP. 

Recently, there has been an increasing interest in automatically designing MOEAs~\cite{Radulescu2013,jmetal_irace,moead_irace}. In these approaches, a configurator recombines components of established MOEAs, creating more effective variants of the original MOEA. However, why these variants perform well is only briefly analysed. We argue that by understanding why these choices of components for multiobjective algorithms are made, we can develop new and better components.

Here, we focus on analysing an instance of Multiobjective Evolutionary Algorithm Based on Decomposition (MOEA/D) ~\cite{zhang2007moea} from a component-wise point of view in MOPs with complicated constraints. The MOEA/D is a popular and efficient algorithm for solving MOPs and can modify its behaviour by considering the constraints of these MOPs. However, to be best of our knowledge, no previous work has analysed how the MOEA/D components affect the dynamics of the search progress of this metaheuristic, especially in constrained MOPs. 

% For analysing the contributions of the selected components, we first automatically search for a well-performing configuration of the MOEA we are studying, considering the contribution of individual components of this optimiser, according to the characteristics of the constrained MOPs. We call this final configuration auto-MOEA/D. Then, we conduct an experiment in which the auto-MOEA/D is compared against its variants, with at most one single component altered, following a similar design used Campelo and Wanner's work~\cite{caiser_2020}.

To analyse how the components contribute to an automatically designed MOEA, first, we automatically generate a well-performing algorithm instance using irace on a set of constrained MOPs. Then we conduct an experiment where the designed algorithm is compared against a set of variants, each of which has at most a single component altered. The experiment design to compare the variants follows the work of Campelo and Wanner~\cite{caiser_2020}.

We obtain variants of the automatically designed MOEA in one of these two ways: (1) when we can remove the component, the variant algorithm takes the structure of the machine-generated MOEA minus this component; (2) otherwise, we replace this component with its counterpart from the original algorithm. In this way, we can find the most influential components of the automatically designed MOEA configuration for the problems studied here. Moreover, we can also verify how these influential components affect the algorithm's ability to navigate the landscape of MOPs here studied.

%% details or this proposal
This investigation takes the form of a case study on nine continuous benchmark problems with two and three objectives, with tunable constraints. We conduct our analysis focusing on how these metaheuristics explore the decision space. Furthermore, we contrast automatically designed MOEA against each of the variants in terms of their Search Trajectory Networks (STNs)~\cite{stn_main,STN_MOP_evostar}; the diversity of the population; and the traditional metrics of hypervolume and inverted generational distance (IGD). To the best of our knowledge, this is the first component-wise analysis of MOEAs about the decision space dynamics in constrained MOPs. For reproducibility purposes, all the code and experimental scripts are available online at \href{https://github.com/yurilavinas/MOEADr/tree/gecco22/}{https://github.com/yurilavinas/MOEADr/tree/gecco22/}.

The paper is organised as follows. Section~\ref{related-work} overviews previous work related to the automated design of algorithms and constrained problems. Section~\ref{prelim} introduces relevant concepts. The automatic design of the MOEA is shown in Section~\ref{exp_automatic}. Then, the comparison of the components setup is presented in Section~\ref{search_behavior}, and the analysis of the search behaviours dynamics of the different MOEA/D variants is shown in Section~\ref{dynamics}. Finally, Section~\ref{discussion} outlines our main findings, limitations and suggestions for future work.

\section{Related Work}\label{related-work}

Most approaches on the automatic design of evolutionary algorithms focus on creating templates that can instantiate many algorithms and their parameter settings for performance improvements. For example, there have been studies to automatically design NSGA-II~\cite{jmetal_irace} and MOEA/D~\cite{moead_irace} on commonly used benchmark sets. Moreover, two seminal examples are the works of Bezerra et al.~\cite{bezerra2016automatic,Bezerra2020}, which proposed a component-wise MOEA template that instantiates multiple existing frameworks for continuous and combinatorial optimisation MOPs. 

Their research efforts mainly focus on exploiting the automatic configuration to increase the performance of multiobjective algorithms in benchmark problems without constraints. We also highlight the work of Radulescu et al.~\cite{Radulescu2013}, that focus on improving the performance of multiobjective metaheuristics. These works are insightful visual approaches; however, they concentrate on finding well-performing configurations of multiobjective algorithms. On the other hand, there are few studies in the context of the automatic design of algorithms focusing on \emph{the effect of the different components} on the algorithm's performance.

Multiple works provide visualisations of Pareto front approximations for the analysis of the MOP algorithms to help in understanding and contrasting algorithm behaviour~\cite{performance_assessment,KerschkeG17,OnePLOTtoShow,Liefooghe2018OnPL,Fieldsend2019VisualisingTL}. However, all of these works focus on increments in performance and analyse the effects of the search progress with respect to the objective space, with little focus on \emph{the decision space dynamics}. We argue that analysing the decision space might expand our understanding of the behaviour of the multiobjective optimisation solvers. That is why we are focusing here on studying and identifying useful metrics that evaluate these metaheuristics from the perspective of exploring the decision space dynamics. 

We highlight a recent work~\cite{STN_MOP_evostar} that generalises to multiobjective optimisation a recent graph-based modelling tool, Search Trajectory Networks (STNs) \cite{stn_evostar,stn_main}. These STN model the search behaviour of MOEAs, allowing us to extract features and information from regions of the decision space that an algorithm has transversed. Therefore, we use these STN models as one of our tools to discriminate the behavioural differences of MOEAs.
\section{Preliminaries}
%% technical
\label{prelim}

% \subsection{MOEA/D}

% The evolutionary computation community has proposed several multi-objective evolutionary algorithms (MOEAs) which can be classified into three broad categories, based on dominance \cite{deb2002fast}, indicators \cite{beume2007sms} and decomposition \cite{zhang2007moea}. They can modify their behaviour when searching for solutions according to the MOP in question, taking into account the constraints of each of these problems. These problems involve multiple conflicting objectives, and finding good sets of solutions for such problems is generally considered a hard problem. 

% Few studies considered the contribution of individual components of MOEAs to performance~\cite{bezerra2016automatic}. In addition to this, few of these algorithms have their performance evaluated in unconstrained MOPs, or the constraints in the MOPs studied are simple to address~\cite{TANABE2020,bezerra_ecj,felipe_CHT}. These MOPs usually have two main characteristics: (1) an unknown and irregular shape of the Pareto front and (2) a highly unfeasible (constrained) objective space. These constraints invalidate some solutions, which makes finding a set of feasible solutions a challenging task.

Few studies have considered the contribution of individual components to MOEAs performance~\cite{bezerra2016automatic}. Furthermore, in most cases, the performance is evaluated in unconstrained problems or in problems where the constraints are simple to address~\cite{TANABE2020,bezerra_ecj,felipe_CHT}. These constraints invalidate some solutions, which makes finding a set of feasible solutions a challenging task.

The MOEA/D is a popular and efficient algorithm for finding good sets of trade-off solutions for MOPs. The key idea of MOEA/D is to create a linear decomposition of the MOP into a set of single-objective subproblems. Decomposing the MOP in various single-objective subproblems makes the algorithm very flexible for dealing with constraints because adding a penalty value is straightforward: MOEA/D adds a penalty value related to the amount of violation of the constraint for each one of the subproblems. Given this nature of the single-objective subproblems, MOEA/D can easily use multiple constraint handling techniques (CHTs).

% I still feel that this algorithm is not very useful in the paper right now. The current position of the algorithm is near the part where you describe MOEA/D for the first time. So maybe change the algorithm so that it becomes more conceptual, to describe the general flow of MOEA/D? People who want details can go to the code.

\begin{algorithm}[t]
	\caption{MOEA/D outline}\label{algo:moead}
	\begin{algorithmic}[1]
		\State Initialize population and decomposition vectors.
		
		\While{\textit{Computational budget is not meet}}
		    
            \State \textbf{Define} Neighbourhood relations.
		    
		    \State \textbf{If} partial update is used, select subset of solutions to update. 
		    \State \textbf{Else}, all solutions are updated.
            
            \State \textbf{Generate} candidates from the updated solutions and their neighbors.
            \State \textbf{Evaluate} the candidates on their respective subproblems.
            
            \State \textbf{Update} the population using the candidates.
            \State \textbf{Save} non-dominated solutions in UEA.
            
		    \State \textbf{if} restart criteria is met, re-generate the population.
		\EndWhile
	\end{algorithmic}
\end{algorithm}

The MOEA/D template we propose for instantiating and
designing variants of this metaheuristic is shown in Algorithm~\ref{algo:moead}. We use the generational version of MOEA/D incremented with the Unbounded External Archive (UEA). The UEA is used to keep all nondominated solutions found by a multi-objective optimizer during the search process. Solutions in the archive are only used as the output of the algorithm and are stored in a way that they do not affect the search run~\cite{uea, uea2}.

\subsection{Automatic Design Configurator}

For the automated design, we use Iterated Racing (irace)~\cite{LOPEZIBANEZ201643}. The irace package searches for a well performing set of components of an algorithm over a set of optimization problems. After fine-tuning the MOEA/D with irace, we conduct an ablation analysis~\cite{fawcett2016analysing, LOPEZIBANEZ201643} to help us understand the choice of components values and whether each of these choices effectively improves the MOEA/D performance. This analysis investigates differences between configurations. We conduct an ablation analysis between the first configuration tried and the best configuration found by irace.

\subsection{Search Trajectory Networks (STNs)}

To visually and quantitatively analyze the dynamics of different algorithm variants, we use the recently introduced extension of search trajectory networks (STNs)\cite{stn_main} for MOPs~\cite{STN_MOP_evostar}.

% comment about the feasibility
In an STN model, each solution in the search space is mapped to a location. Similar solutions are generally mapped to the same location, as the locations represent a partition of the search space. The network models are extracted from data obtained during several runs of the studied algorithm(s). A network model requires defining its nodes and edges. In an STN model,  \textit{nodes} are locations in the search trajectories visited by a given algorithm, an \textit{edges} connect two consecutive locations in the trajectory. A strength of the network models is that they can be visualized. When decorating the networks for visualization, it is possible to highlight attributes of the nodes and edges that are relevant to the search process. In these visualizations, the size of nodes and the width of edges are proportional to how many times they were visited by the algorithms during the aggregation of runs used to extract the model. The color of nodes and edges also highlight relevant attributes as indicated in the legends of Figures \ref{fig:stn_3components_DASCMOP1} and \ref{fig:stn_3components_DASCMOP6}.  Visualisations use \emph{force-directed} graph layout algorithms as implemented in \textsf{R} package igraph \cite{igraph}.

Here, we extend this approach by using the idea of merging the trajectories of two or more single-objective algorithms proposed in \cite{stn_main}. Thus, we combined the STN models of different MOEA/D variants into one single merged STN model. This allows us to directly visually compare how distinct variants explore the search space. In the merged models, there is the notion of \textit{shared nodes}, which are nodes visited by more than one algorithm and are indicated in grey colour in the network visualization.

\subsection{Network and Performance Metrics}

We use three STN metrics to assess the global structure of the trajectories and bring insight into the behaviour of the MOEAs modelled. These metrics are (1) the number of unique nodes and (2) the number of unique edges, and (3) the number of shared nodes between vectors. These metrics are summarised in Table~\ref{stn_metrics}. We also use the population variance metric.

For reference, we use the following criteria to compare the results of the different strategies: (a) final approximation hypervolume (HV), the volume of the n-dimensional polygon formed by reference points and solutions and (2) inverted generational distance (IGD), the mean distance between reference points to the nearest solutions. It is worth noting that additional decision space and MOP metrics could also be considered.

\begin{table}[htbp]
\centering
\caption{Description of decision space metrics}
\label{stn_metrics}
\begin{tabular}{c|c}
\rowcolor[gray]{.85}Metric & Description \\ 
    Nodes & Unique locations visited.\\
    
    \multirow{2}{*}{Edges} &          Unique search transitions \\
    & between nodes.\\
    
    \multirow{2}{*}{Variance} &          Dispersion is the population \\
    & in the decision space.\\
    
    % Variance & Dispersion is the population in the decision space.\\
\end{tabular}
\vspace{-0.5em}
\end{table}

\section{Designing auto-MOEA/D}
\label{exp_automatic}

In this work, we analyse the components of a MOEA/D instance that were automatically designed. This design process was done in a component-wise framework, similar to the protocols used by Bezerra et al.~\cite{bezerra2016automatic} and Campelo et al. ~\cite{moeadr_paper}. We extend the MOEADr package~\cite{moeadr_package} to introduce options for population restart and partial update of the population ~\cite{lavinas2020moea,lavinas2022faster}.

\subsection{Variable Components Search Space}

The configuration search space used in our experiments contains the algorithm components and numerical parameters of the \\MOEA/D framework. These are shown in Table~\ref{automated_parameter_table}.

Special attention is required with the variation operators: Differential Evolution (DE) mutation and polynomial mutation. They are always performed sequentially, first DE and then the polynomial mutation. Thus, the order of the stack of operators is kept fixed, but the parameter values are variable. Similar attention should be given to the restart strategy, where only the strategy's choice is explored. 

\begin{table}[t]

\centering
	\small
	\caption{Components search space.}
	\label{automated_parameter_table}
    \begin{tabular}{l|l}
        % \hline
        
        \rowcolor[gray]{.85}Component            & Domain            \\ 
        \multirow{2}{*}{Decomposition vector generator}    & Uniform or \\ 
                                                        & Sobol \\ \hline
        \multirow{2}{*}{Population size}         & 100 or \\ 
                                                 & 500     \\ \hline    
        \multirow{2}{*}{Aggregation function}         & Weighted Tchebychef or\\
                                                      & Adjusted weighted Tchebychef\\ \hline    
        \multirow{2}{*}{Update strategy}                & Best, $nr = [1, 20]$   \\
                                                        & Restricted, $Tr = [4, 20]$  \\ \hline    
        \multirow{2}{*}{Neighbourhood function} & $T  = [10, 100]$\\    
                                                      & $Delta = [0.1, 1]$ \\ \hline   
        DE mutation                                  & $F = [0.1, 1]$       \\ \hline
        \multirow{2}{*}{Polynomial mutation}        & $\eta_m = [1, 100]$    \\
                                                            & $prob = [0, 1]$     \\ \hline
        \multirow{2}{*}{Partial update strategy}        & True $n = {0.10, 0.15, 0.20, 0.25}$ \\
                                                        & False, not used\\ \hline  
        \multirow{2}{*}{Restart strategy}        & True, every $20000$ evaluations  \\
                                        & False, not used 
        \end{tabular}
        % \vspace{-1.5em}
\end{table}

% \subsection{Fixed Parameters}

We choose to fix some MOEA/D components to reduce the search space for the irace configurator. The fixed components are the computational budget, the objective scaling and the constraint handling technique (CHT). These fixed components are always present in every configuration of the MOEA/D that irace generates: (1) the number of functions evaluations is set to $100000$ in order to grasp all possible behaviours of the automatically designed algorithm during the run; (2) all objectives were linearly scaled at every iteration to the interval $\left[0,1\right]$; (3) we use the Dynamic CHT~\cite{joines1994use}, which starts with a small penalty value, increases it across the iterations to focus on the diversity of feasible solutions, and then later focus on the convergence of those solutions. It is defined by $f^{agg}_{penalty}(x)  = f^{agg} + (C*t)^\alpha*v(x)$, where $C = 0.05$ and $\alpha = 2$ are constants we defined based on the following works~\cite{joines1994use,felipe_CHT}.

% \begin{equation}
%       f^{agg}_{penalty}(x)  = f^{agg} + (C*t)^\alpha*v(x)
% \end{equation}

% where $C = 0.05$ and $\alpha = 2$ are constants we defined based on the following works~\cite{joines1994use,felipe_CHT}.

\subsection{Configurator Setup}

We use the irace configurator~\cite{LOPEZIBANEZ201643} to automatically assemble and design a MOEA/D configuration based on the components available in the MOEADr package extended with the options for population handling mentioned above. We run irace with its default settings, except for the number of elite configurations tested, that we increase from 1 to 7, following Campelo et al. work~\cite{moeadr_paper}. We run irace with a budget of $15 000$ runs.

\subsection{Benchmark Problems}

We use the DASCMOP benchmark set~\cite{dascmop}. This set has nine constrained test functions: DASCMOP1-6, each with eleven constraints, and DASCMOP7-9, each with seven constraints. The constraints can be modified to consider three types of difficulties: type-I considers diversity-hardness, type-II considers feasibility-hardness and type-III consider convergence-hardness. More information about the problems and these difficulty triplets can be found in~\cite{dascmop}. We use the implementation of the test problems available from the \textit{Pymoo} python package~\cite{pymoo}. Since our work is focused on  MOPs with complicated constraints, we select one of the most challenging difficulty triplets, numbered $16$, which considers all three types of constraints difficulties.

\subsection{Evaluation Metrics}

Analysing MOP solvers considering only their final approximation provides limited information related to these algorithms' performance since any MOP solver should return a suitable set of solutions at any time during the search \cite{zilberstein1996using,radulescu2013automatically,anytimePLS,tanabe2017benchmarking}. Here, we analyse the anytime performance effects in terms of hypervolume (HV) values to investigate the impact of different configurations of MOEA/D on their Unbounded External Archive (UEA)~\cite{uea}.

We use the following method to compare the results of the different strategies: we calculate the accumulative HV over the search progress to quantify the HV anytime performance. At every $1000$ evaluations, we calculate the HV of the solutions in the UEA at that iteration, using the reference point as (11, over the number of objectives), following the work of Bezerra et al.~\cite{bezerra2016automatic}. Then, we sum all values to have an approximated evaluation of the anytime HV curve.

% \subsection{Configurator Results}\label{trends_auto}

% % \begin{figure*}[htbp]
% % \centering
% % \includegraphics[width=1\textwidth]{imgs/parameters-1.pdf}
% % \caption{irace output with the frequency of the different choice of components and parameters.}
% % \label{fig:stn_irace_output}
% % \end{figure*}

% % Figure~\ref{fig:stn_irace_output} shows the frequency of the different choice of components and parameters after the tuning and ablation design is performed. There is a consensus over almost all the components and parameters studied here for some components or parameters, except for the update strategy and restart strategy.

Here, we briefly describe the performance of auto-MOEA/D in terms of HV~\footnote{higher is better.} and inverted generational distance (IGD)~\footnote{lower is better.} and variance of the population, and the STNs metrics: number of nodes, edges and shared nodes. Since we use the reference point of $11$ over the number of objectives, the maximum HV is $121$ for MOPs with two objectives or $1331$ for MOPs with three objectives.

\begin{table}[htbp]
\centering
% \vspace{-1em}
\caption{\textit{hypervolume (HV)},  \textit{IGD},  \textit{Nodes},  Edges and population \textit{variance} of the auto-MOEA/D.}
\label{metrics_ablation}
\begin{tabular}{c|ccccc}
    % \hline
    \rowcolor[gray]{.75} MOP &\multicolumn{5}{c}{\textbf{auto-MOEA/D}} \\
    \rowcolor[gray]{.82} & HV/max(HV) & IGD & Nodes &  Edges & Var.\\ 
    DASCMOP1 & 0.27 & 0.23 & 8469 & 10137 & 0.06 \\
    DASCMOP2 & 0.58 & 0.27 & 8601 & 10291 & 0.03 \\
    DASCMOP3 & 0.52 & 0.32 & 11723 & 13636 & 0.05 \\
    \rowcolor[gray]{.95}DASCMOP4 & 0.01 & 2.72 & 2140 & 2687 & 0.16  \\
    \rowcolor[gray]{.95}DASCMOP5 & 0.00 & 5.23 & 2291 & 2687  & 0.36  \\
    \rowcolor[gray]{.95}DASCMOP6 & 0.09 & 4.03 & 2502 & 3246 & 0.57  \\
    \rowcolor[gray]{.95}DASCMOP7 & 0.01 & 5.17 & 2437 & 3373 & 0.12 \\
    \rowcolor[gray]{.95}DASCMOP8 & 0.00 & 5.03 & 2277 & 3029  & 0.09 \\
    DASCMOP9 & 0.14 & 0.57 & 12952 & 16285 & 0.12
\end{tabular}
% \vspace{-1em}
\end{table}

The different metrics values for the auto-MOEA/D are shown in Table~\ref{metrics_ablation}. We can see that higher HV values and lower IGD values correspond to a high number of nodes and generally low populational variance values in the DASCMOP1, DASCMOP2, DASCMOP3 and DASCMOP9 problems. In contrast, the opposite happens for the other problems (highlighted in light grey).

Thus, we separate these problems into two groups: the easy group, with DASCMOP1-3 and DASCMOP9, in which MOEA/D performs well; and the hard group, with DASCMOP4-6, in which MOEA/D performs poorly. This separation is in agreement with the work of Fan et al.~\cite{dascmop}, where they state that DASCMOP1-3 and DASCMOP4-6 have similar feasible regions, distance function and Pareto Fronts when they have the identical difficulty triplets (number $16$ in our case).
\section{Comparison of the Components}\label{search_behavior}

\begin{figure*}[htbp]
\centering
% \vspace{-1.5em}
	\begin{subfigure}[!t]{0.5\textwidth}
        	\includegraphics[width=1\textwidth]{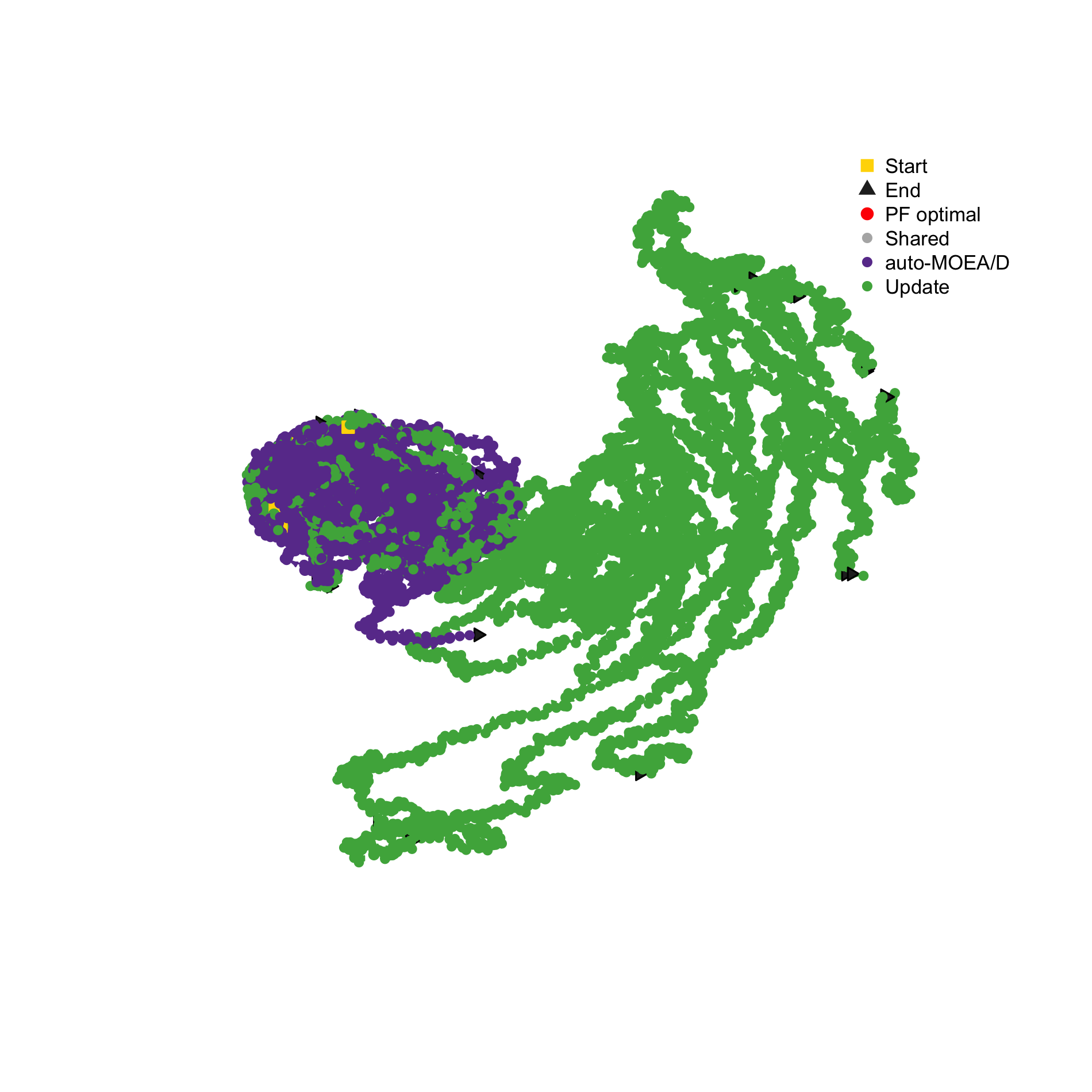}
        % 	\vspace{-4.5em}
        	\caption{DASCMOP1: the trajectories of the two variants overlap.}
	\end{subfigure}
	~~
    \begin{subfigure}[!t]{0.5\textwidth}
        	\includegraphics[width=1\textwidth]{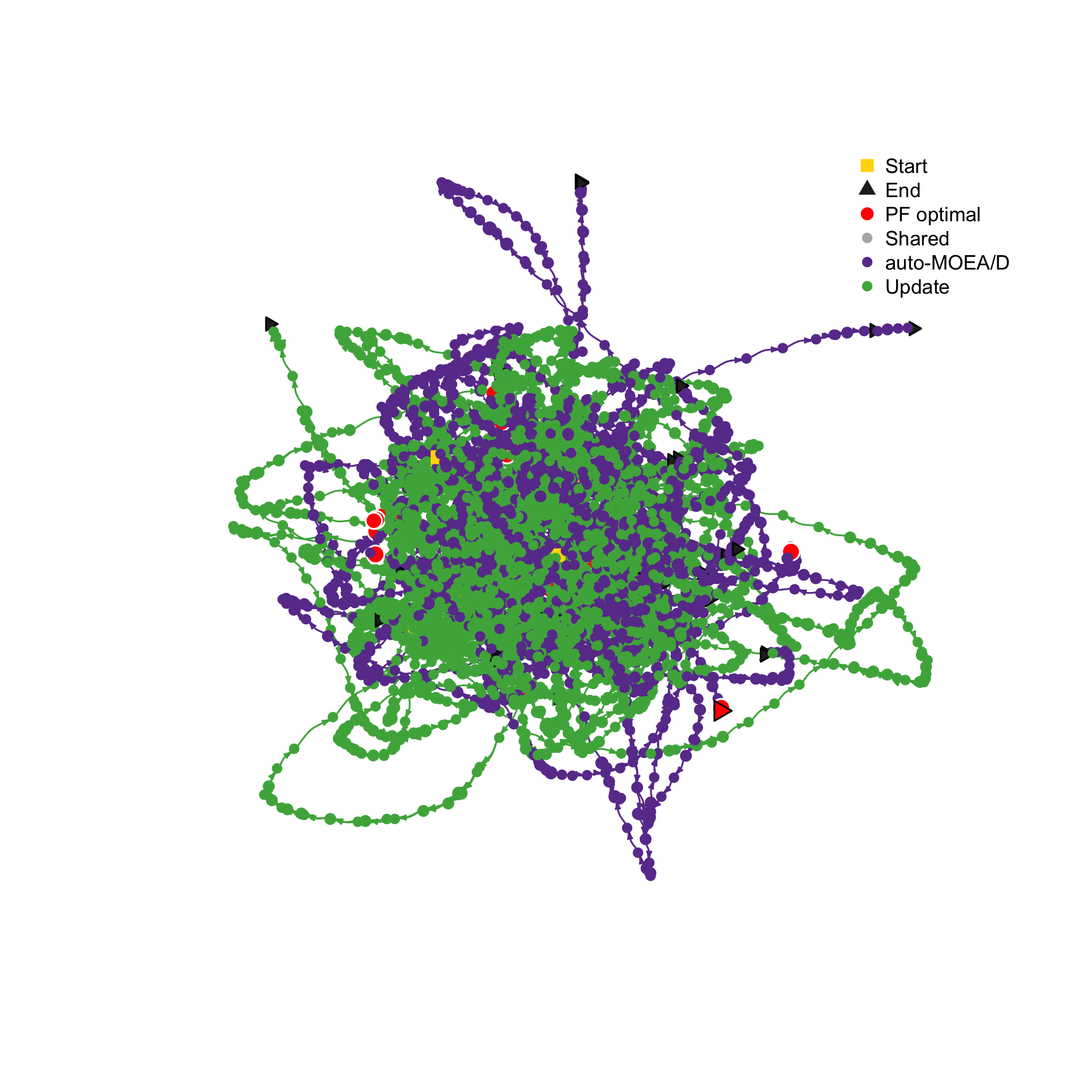}
        % 	\vspace{-4.5em}
        	\caption{DASCMOP6: the trajectories of the two variants do not overlap.}
	\end{subfigure}
\caption{STNs of auto-MOEA/D and the update strategy on two MOP instances. We can see that the variants traverse similar regions in the decision space. On the left, we can see that the update strategy variant explore more areas of the search but no algorithm is able to find optimal solutions, while on the right, the opposite happens.}
\label{fig:stn_3components_DASCMOP1}
\end{figure*}

\begin{figure*}[htbp]
\centering
% \vspace{-1.5em}
	\begin{subfigure}[!t]{0.5\textwidth}
        	\includegraphics[width=1\textwidth]{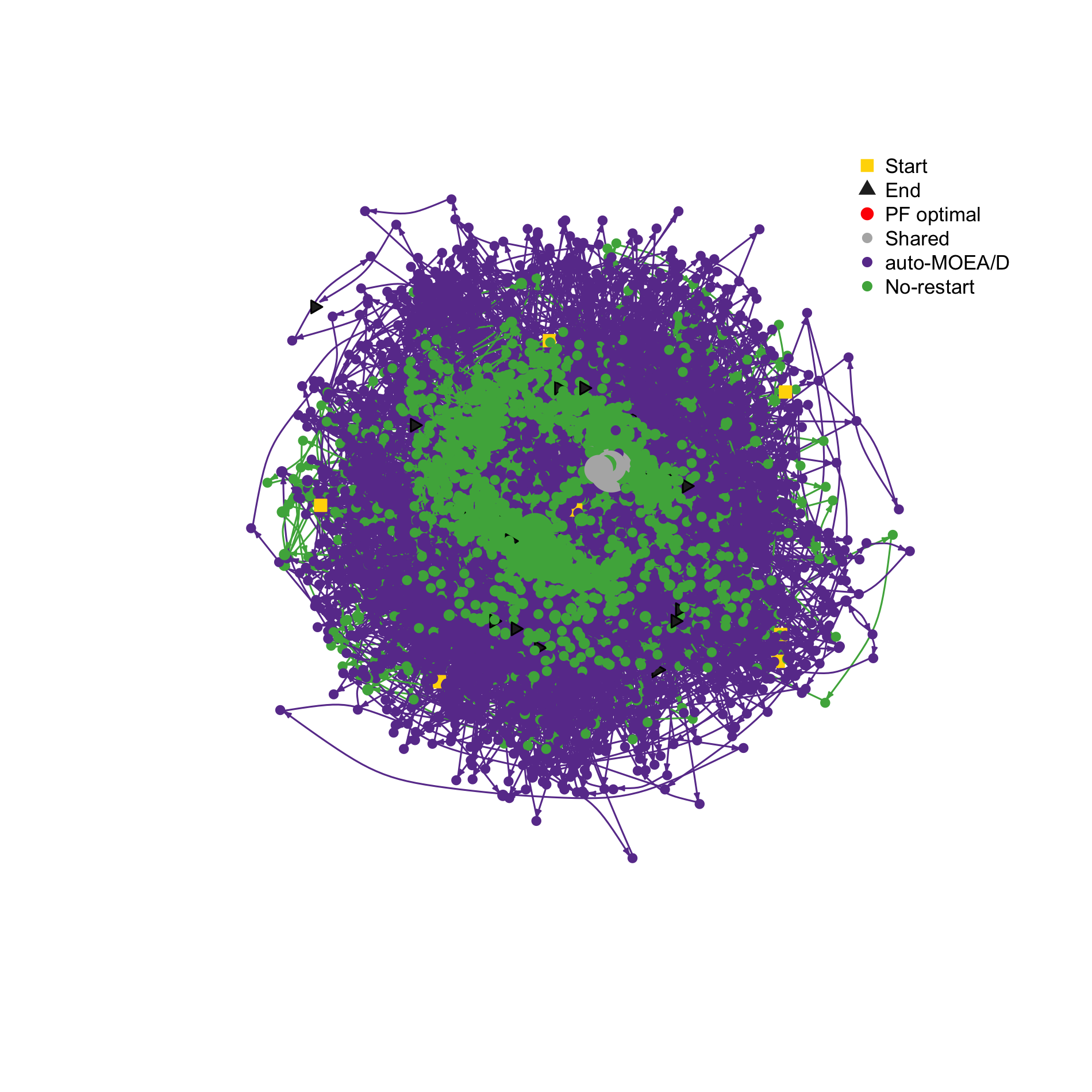}
        % 	\vspace{-4.5em}
        	\caption{DASCMOP1: the trajectories of the two variants overlap.}
	\end{subfigure}
	~~
    \begin{subfigure}[!t]{0.5\textwidth}
        	\includegraphics[width=1\textwidth]{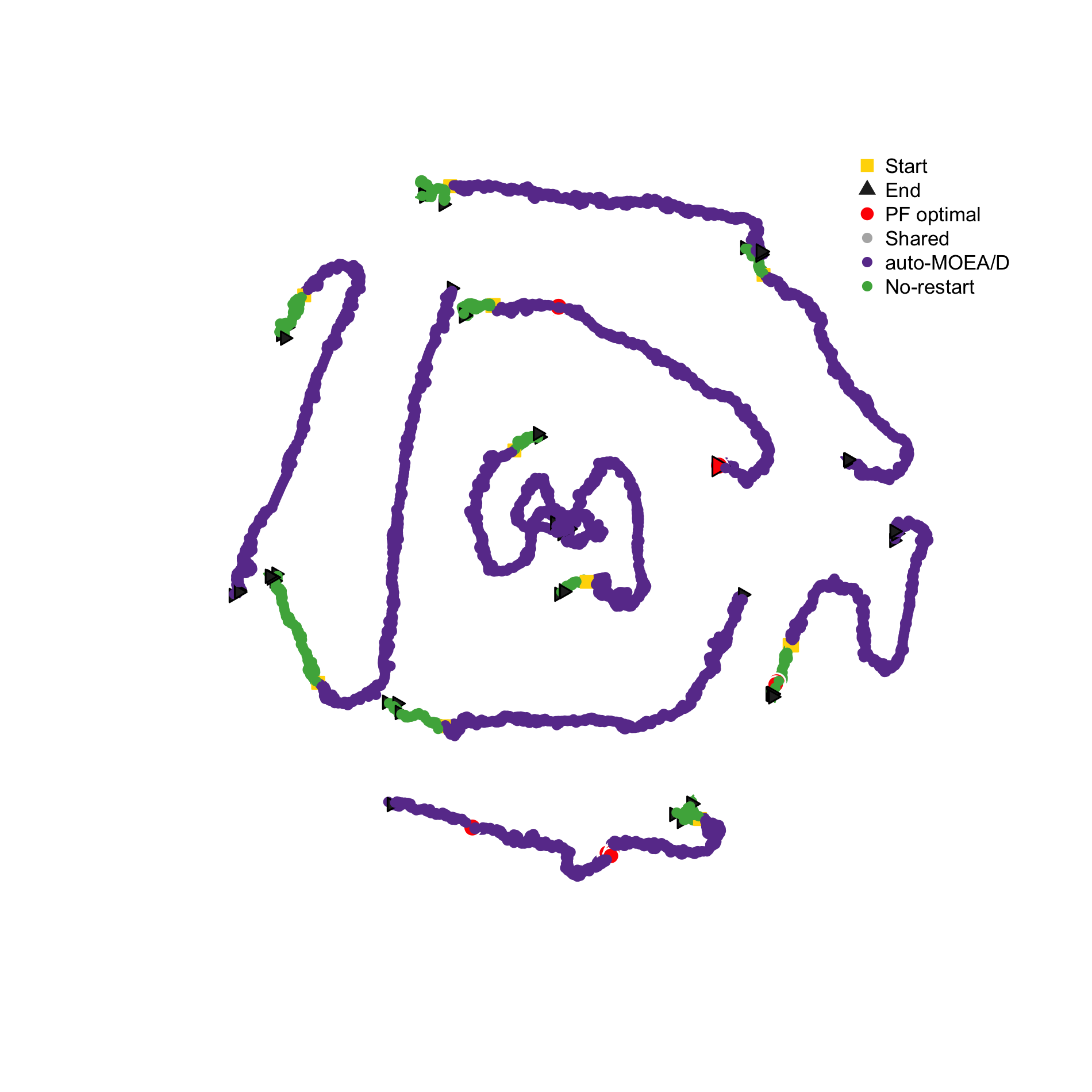}
        % 	\vspace{-4.5em}
        	\caption{DASCMOP6: the trajectories of the two variants do not overlap.}
	\end{subfigure}
\caption{STNs of auto-MOEA/D and the no-restart strategy on two MOP instances. On the left, we can see that the variants traverse similar regions in the decision space visiting some of the nodes multiple times, while on the right, the opposite happens.}
\label{fig:stn_3components_DASCMOP6}
\end{figure*}

Here we introduce a methodology for investigating the effects of the final configuration of a machine-designed multiobjective algorithm. This analysis aims to measure the differences in the search dynamics among several variants from the MOEA and, through these measures, identify the most influential components of the automatically designed algorithm.

\begin{table}[htbp]
\caption{Auto-MOEA/D setup in this work, and the variants under analysis. For each variant, only \textit{one component} is changed, while the other components are the same as auto-MOEA/D as shown below.}
\label{variants}
\begin{tabular}{c|c}
\rowcolor[gray]{.75}\multicolumn{2}{c}{\textbf{Auto-MOEA/D setup, and its Variants}}\\ 
\rowcolor[gray]{.85}\textbf{auto-MOEA/D}                    & \textbf{Component variant} \\ 

\textbf{Decomposition}      &   \textbf{Decomposition} \\       $Uniform$                   & $SLD$                    \\ \hline

\textbf{Population size}      &   \textbf{Population size} \\           $100$    & $300$                       \\ \hline

\textbf{Update}      &   \textbf{Update} \\
$Restricted$    & $Restricted$           \\ 
$nr = 13$        & $nr = 2$         \\  \hline

\textbf{Neighbourhood pars.}      &   \textbf{Neighbourhood pars.} \\
$T = 18$    &  $T = 20$                   \\
$Delta = 0.5831$    &   $Delta = 0.9$       \\ \hline

\textbf{Operators pars.}      &   \textbf{Operators pars.} \\
$DE: F = 0.705$    &  $DE: F = 0.5$                          \\
Polynomial: $\eta_m = 57.0443$    &  Polynomial:  $\eta_m = 20$     \\
Polynomial: $prob = 0.303$    &   Polynomial: $prob =  0.3$   \\ \hline

\textbf{Restart}      &   \textbf{Restart} \\                           $True$    & $False$                           \\ 

\end{tabular}
\end{table}

To analyse these behavioural effects of the different variants, we compare the auto-MOEA/D described in Section~\ref{exp_automatic} against variants with at most one single component altered. We obtain these variants by changing or removing a single component the auto-MOEA/D at a time. 
This is done by either (1) removing the component from the algorithm when possible or (2) changing its parameters to its counterpart in the traditional MOEA/D. Table~\ref{variants} lists these differences. We chose \textit{six} variants to analyse, which added to auto-MOEA/D itself least to a total of \textit{seven} variations of the MOEA/D family. These variations are compared quantitatively and visually in terms of their STN models to detect which components affect the most extensive changes.

\section{Behavioural dynamics}\label{dynamics}

To quantitatively analyse the dynamics of the search progress of the different variants of MOEA/D, we mode the search dynamics using STNs for each pair of auto-MOEA/D and an auto-MOEA/D variant, leading to six different pairs. 

We base our quantitative analysis on the traditional multiobjective metrics hypervolume (HV) and inverted generational distance (IGD) and the number of nodes, edges and shared nodes of the STN models and the populational variance. For the HV, we use the reference point of 1.1 over the number of objectives. We linearly scaled all objectives to the interval $[0, 1]$, for a more straightforward comparison among the algorithms.

\begin{figure}[htbp]
\centering
\includegraphics[width=0.47\textwidth]{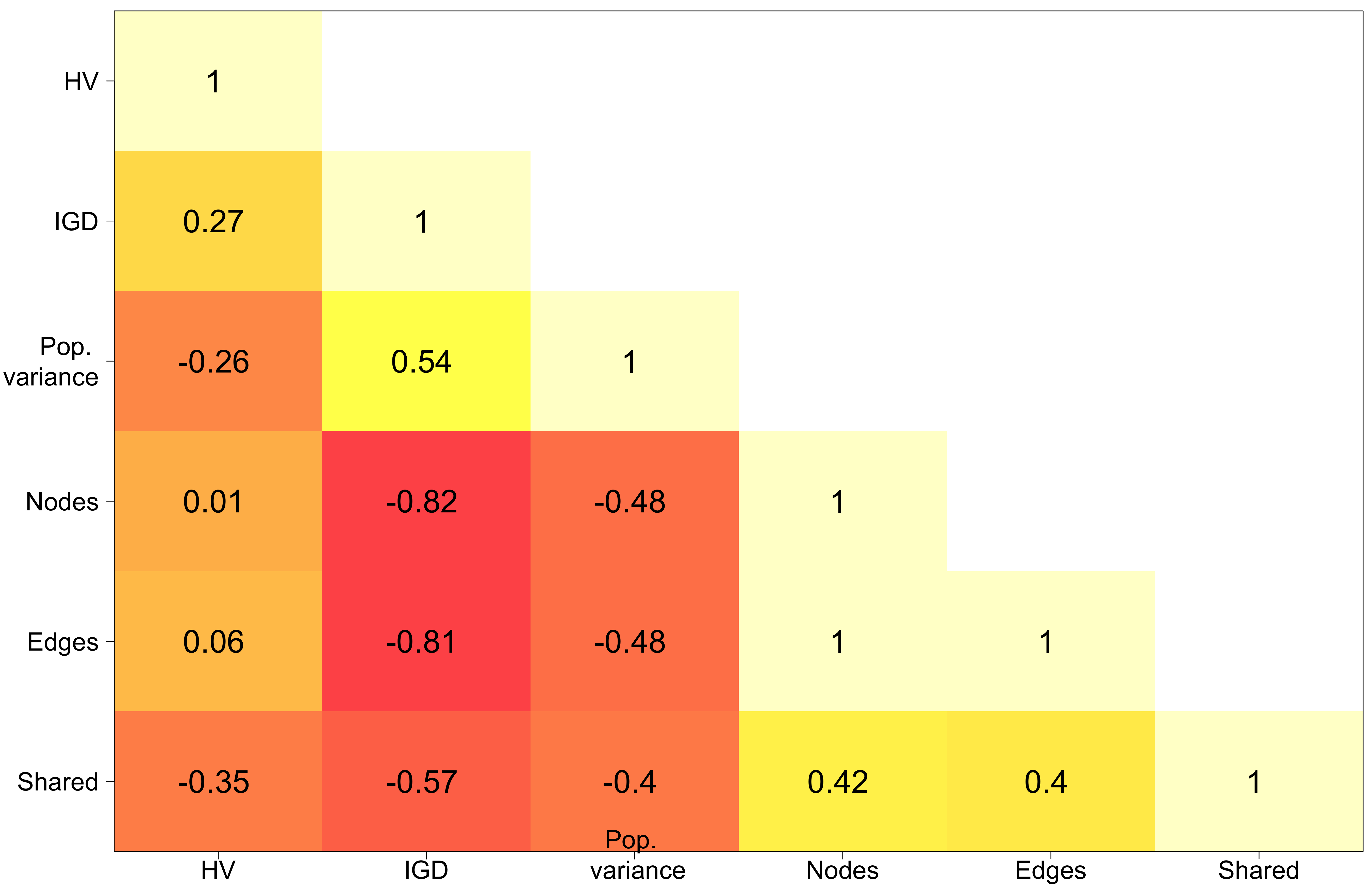}
\caption{Correlation matrix among the different metrics. We find some correlation among HV and shared nodes; and some correlation among IGD, number of nodes and edges.}
\label{fig:correlation_all}
\end{figure}

\subsection{Metrics Analysis}

Figure~\ref{fig:correlation_all} shows the correlation matrix among the different metrics studied, considering the results of the MOEA/D variants in the DASCMOP set of problems. Although the HV and IGD metrics are helpful for objective space analysis, we understand that they provide useful information for the base of the analysis; thus, we keep them in our analysis. 

Since the number of nodes and edges have a high correlation between themselves and among the IGD, we remove the number of edges metric from our analysis. The other STN metrics, together with the population variance are on the decision space analysis; thus, we use them in our following analysis to strengthen the study. Furthermore, we see a correlation among the IGD metric between the number of nodes, edges, and shared nodes and the population variance. This correlation suggests a link between these decision space metrics and IGD increments of performance. 

%%%%%%%%%%%%%%%%%%%%%%%%%%%%%%%%%%%%%%%%%%%%%%%
\subsection{STNs Extension for Pairs of MOEAs}

For creating merged STN models of pairs of MOEAs, we first need to create one STN for each algorithm. To create the STN of a single algorithm, we follow a recently proposed methodology \cite{STN_MOP_evostar}. 

\paragraph{Basic operations.} The STNs for multiobjective optimization rely on the notions of decomposition and scalar aggregation function operations~\cite{zhang2007moea}:

\begin{description}

\item[\bf Decomposition.] Breaks down a MOP into a set of single objective subproblems, where each subproblem is a combination of the objectives, characterised by a weight vector. In this work, we use the Uniform Design~\cite{uniform_design}, as it allows us to choose the number of weight vectors to be generated explicitly.

\item[\bf Scalar aggregation function.] Takes the objective values of one solution and the weight vector of one subproblem and calculates a scalar value that represents the quality of that solution for that particular subproblem. We use the Weighted Tchebycheff~\cite{miettinen1999nonlinear, moeadr_paper}, which is less affected by the shape of the Pareto front in a given MOP.
\end{description}

\paragraph{Basic methodology.} We summarise below the methodology proposed in \cite{STN_MOP_evostar}. 

\begin{enumerate}
    \item Choose the number of decomposition vectors $n$ and generate these vectors using the decomposition technique. In this study we set $n=5$.
    \item At every iteration, select a representative solution for each vector, using the scalar aggregation function. In case of ties, choose the newest solution.
    \item Map each representative solution as the nodes, with a precision parameter to portion the space with length $10^{-01}$.
    \item Edges are given by the sequence of locations of consecutive iterations. 
    \item Each decomposition vector is modelled as an STN trajectory.
    \item Merge these STNs trajectories by the graph union of these graphs.
\end{enumerate}

This approach allows us to create the STNs of single algorithms. As discussed (Section~\ref{prelim}), we extend this approach by merging the trajectories of two of this STNs by joining the two STNs graphs. This merged STN model contains the nodes and edges present in the STN of at least one algorithm. Attributes are kept for the nodes and edges, indicating whether they were visited by both algorithms (shared) or by one of them only. 
%%%%%%%%%%%%%%%%%%%%%%%%%%%%%%%%%%%%%%%%%%%%%%%

\subsection{Search Behaviour Dynamics}

\begin{table*}[htbp]

\centering
% \caption{\textit{$\Delta$HV}; \textit{$\Delta$IGD}; number of \textit{$\Delta$nodes}; \textit{$\Delta$variance} of the population and  number of \textit{Shared} nodes   given the different variants of the auto-MOEA/D in terms of components on the constrained problems. For the first three metrics, positive values indicate an increase in performance in relation to the auto-MOEA/D while negative values indicate a decrease performance. For the number of shared nodes, higher values indicate similarities in the search dynamics.}
\caption{\textit{$\Delta$HV}; \textit{$\Delta$IGD}; number of \textit{$\Delta$nodes}; \textit{$\Delta$variance} of the population and  number of \textit{Shared} nodes   given the different variants of the auto-MOEA/D in terms of components on the constrained problems.}
\label{metrics_all}
\begin{tabular}{c|ccccc|ccccc}
    % \hline
    \rowcolor[gray]{.75}  &\multicolumn{5}{c|}{\textbf{Decomposition variant}} & \multicolumn{5}{c}{\textbf{Population size variant}}\\
    \rowcolor[gray]{.82} & $\Delta$HV & $\Delta$IGD &  $\Delta$nodes  & $\Delta$variance & Shared & $\Delta$HV & $\Delta$IGD &  $\Delta$nodes  & $\Delta$variance  & Shared \\ 
    
    DASCMOP1 &0.02 & -0.01 & 75 & 0.00 & 128
    & -0.08 & 0.03 & 264 & -0.04 & 77
    \\
    
    DASCMOP2 & 0.01 & 0.00 & 253 &0.00 & 122
    & -0.01 & 0.01 & 284 & -0.01 & 90
    \\
    
    DASCMOP3 &0.00 & 0.02 & 306 & -0.02 & 42
    & -0.01 & 0.31 & -1600 &-0.02 & 27
    \\
    
    \rowcolor[gray]{.92}DASCMOP4  & -0.10 & 1.63 & 46 & 0.13 & 11
    & -0.01 & 0.24 & 325 & 0.05 & 2
    \\
    
    \rowcolor[gray]{.92}DASCMOP5 & 0.03 & -0.60 & 49 & 0.03 & 12
    & 0.13 & -2.58 & 425 & -0.06 & 5
    \\
    
    \rowcolor[gray]{.92}DASCMOP6 & -0.01 & 0.20 & 75 & 0.38 & 12
    & 0.08 & -1.27 & 471 & 0.66 & 6
    \\
    
    \rowcolor[gray]{.92}DASCMOP7 & 0.00 & -0.04 & 41 & -0.04 & 2
    & 0.11 & -1.30 & 113 & 0.26 & 5
    
    \\
    
    \rowcolor[gray]{.92}DASCMOP8 &0.07 & -0.93 & 227 & 0.01 & 5
    & -0.01 & 0.17 & 453 & 0.02 & 3
    \\
    
    DASCMOP9 &0.00 & -0.03 & -884 & 0.04 & 3
    & -0.02 & 0.03 & -1350 & 0.01 & 6
    \\  
    
    \multicolumn{9}{c}{}\\
    
    \rowcolor[gray]{.75} MOP &\multicolumn{5}{c|}{\textbf{Update variant}} & \multicolumn{5}{c}{\textbf{Neighbourhood parameter variant}}\\
   \rowcolor[gray]{.82} & $\Delta$HV & $\Delta$IGD &  $\Delta$nodes  & $\Delta$variance & Shared & $\Delta$HV & $\Delta$IGD &  $\Delta$nodes  & $\Delta$variance  & Shared \\ 
    
    DASCMOP1 & -0.08 & 0.01 & -82 & -0.03 & 62
    & 0.01 & 0.01 & 116 & -0.03  & 116
    \\
    
    DASCMOP2 & -0.01 & -0.04 & -246 & -0.01 & 68
    & 0.00 & -0.02 & 66 & -0.01  & 109
    \\
    
    DASCMOP3 & 0.00 & -0.02 & -2216 & 0.04 & 35
    & 0.00 & 0.03 & -28 & 0.05  & 48
    \\
    
    \rowcolor[gray]{.92}DASCMOP4 & 0.11 & -1.82 & 1196 & 0.16 & 11
    & -0.08 & -0.08 & 100 & -0.09  & 11
    \\
    
    \rowcolor[gray]{.92}DASCMOP5 & 0.25 & -4.26 & 1091 & 0.14 & 12
    & 0.05 & -0.91 & -75 & 0.24  & 12
    \\
    
    \rowcolor[gray]{.92}DASCMOP6 & 0.20 & -3.26 & 1026 & -0.43 & 13
    & 0.02 & 0.08 & 10 & -0.22  & 12
    \\
    
    \rowcolor[gray]{.92}DASCMOP7 & 0.26 & -3.50 & 1006 & 0.32 & 10
    & -0.09 & 1.38 & 101 & 0.25  & 10
    \\
    
    \rowcolor[gray]{.92}DASCMOP8 & 0.31 & -3.77 & 1346 & 0.56 & 13
    &0.02 & -0.48 & 37 & 0.05  & 13
    \\
    
    DASCMOP9 & -0.05 & -0.01 & -1427 & 0.16 & 19
    &  0.03 & -0.05 & -643 & 0.00  & 21
    \\ 
    
    \multicolumn{9}{c}{}\\
    
    \rowcolor[gray]{.75} MOP &\multicolumn{5}{c|}{\textbf{Operators variant}} & \multicolumn{5}{c}{\textbf{No-restart variant}}\\
    \rowcolor[gray]{.82} & $\Delta$HV & $\Delta$IGD &  $\Delta$nodes  & $\Delta$variance & Shared & $\Delta$HV & $\Delta$IGD &  $\Delta$nodes  & $\Delta$variance  & Shared \\ 
    
    DASCMOP1 & 0.09 & -0.06 & 1994 & 0.05  & 139
    & 0.20 & -0.08 & -3836 & 0.89 & 129
    \\
    
    DASCMOP2 &  0.04 & 0.01 & 1879  & 0.08  & 139
    & 0.12 & -0.01  & -2202  & 0.53 & 117
    \\
    
    DASCMOP3 & -0.01 & 0.00 & 1606 & 0.01  & 55
    & 0.03 & 0.03&  -9962 & 0.15 & 43
    \\
    
    \rowcolor[gray]{.92}DASCMOP4 & -0.05 & 0.27 & -224 & 0.05  & 11
    & -0.24 & 4.22 & -1942 & -0.11 & 11
    \\
    
    \rowcolor[gray]{.92}DASCMOP5 &0.03 & -1.04 & -314 & 0.35  & 12
    & -0.08 & 1.36 & -2056 & -0.27 & 12
    \\
    
    \rowcolor[gray]{.92}DASCMOP6 &0.05 & -1.27 & -323 & 0.39  & 13
    & -0.13 & 2.43 & -2178  & -0.5 & 12
    \\
    
    \rowcolor[gray]{.92}DASCMOP7 &0.00 & -0.09 & -434 & 0.39  & 10
    & -0.22 & 3.67& -2010 & -0.03 & 10
    \\
    
    \rowcolor[gray]{.92}DASCMOP8 &0.01 & -0.56 & 225 & 0.57  & 13
    & -0.19 & 2.99 & -1948 & -0.01 & 13
    \\
    
    DASCMOP9 &0.01 & -0.02 & 60 & 0.11  & 19
    & 0.02 & 0.03 & -7303 & -0.06 & 20
    \\

\end{tabular}
\end{table*}

Based on the results shown above, we compared the auto-MOEA/D and its variants in terms of HV, IGD, population variance, and the STNs metrics: number of nodes and shared nodes. For a more straightforward comparative analysis, we calculate the difference between the results found by the variants to the results found by the auto-MOEA/D (Table~\ref{metrics_ablation}). We show the $\Delta$HV, $\Delta$IGD, $\Delta$nodes, $\Delta$variance. For all of these metrics, positive values indicate larger values in relation to the base algorithm, while negative values indicate the opposite. The only exception in for the IGD values, where negative values indicate larger values in relation to the base algorithm. The number of shared nodes is the only absolute metric.

The different metrics for the variants are shown in Table~\ref{metrics_all}. In terms of the traditional MOP metric HV and IGD, we can see most variants perform the same as the base algorithm in DASCMOP1, DASCMOP2, DASCMOP3 and DASCMOP9, MOPs that compose the easy group (as discussed in Section~\ref{exp_automatic}). The only exception is the no-restart variant that generally leads to more positive differences in the metrics values: high differences in HV, number of nodes, populational variance (except for DASCMOP9); and a good amount of shared nodes. These results suggest that not using the restart strategy can enhance the search ability of MOEA/D in easy problems, leading to HV performance increments, more regions of the search being visited, and more variety in the population. Interestingly, changing the restart strategy has little effect on the number of shared nodes and the IGD values. For the set of hard problems, the performance of the no-restart variant deteriorates in terms of HV, the number of nodes, and population variance, indicating that the restart strategy should be avoided in hard problems but used in easy problems. 

For the other problems (shown in light grey in Table~\ref{metrics_all}), there are more changes in HV and IGD performance, and in this case, the update variant increases the performance the most. For this group of hard MOPs (Section~\ref{exp_automatic}), the update variant generally leads to more differences in the metrics values: positive differences in HV in the number of nodes and negative differences in the IGD values. The populational variance and the number of shared nodes are similar to the other variants. This result suggests that the update strategy has a decisive role in the search ability of MOEA/D in hard problems. Contrary to our expectations, the decomposition variant shows minor differences in all metrics we analysed here. The other variants also show minor differences in these metrics.

Moving on to the STNs visualisations, Figures~\ref{fig:stn_3components_DASCMOP1} and ~\ref{fig:stn_3components_DASCMOP6}. Within the two MOP groups, we see minor changes in the STNs visualisations; therefore, we selected visualisations of a representative problem of each group to show here: from the easy problem, we select the DASCMOP1, and for the hard group, we select the DASCMOP6. Considering the colours used in the STN visualisations, yellow squares indicate the start of trajectories, and black triangles indicate the end of trajectories. The red colour shows the best Pareto optimal solutions, and light grey circles show shared locations visited by both algorithms in that MOP. Finally, the trajectories of each algorithm are shown in different colours: purple for the auto-MOEA/D and green for the variant.

Looking at the images of the STNs of auto-MOEA/D and the update strategy, in the DASCMOP1 and DASCMOP6, Figure~\ref{fig:stn_3components_DASCMOP1}, and the STNs of auto-MOEA/D and the no-restart strategy, also in DASCMOP1 and DASCMOP6, Figure~\ref{fig:stn_3components_DASCMOP6}. For the DASCMOP1, the trajectories of the STNs of the auto-MOEA/D and each variant overlap; that is, they visit similar regions in the decision space. We associate this behaviour with the number of shared nodes for this problem is high: 63 in the case of the update variant pair and 48 in the case of the no-restart variant pair (See Table~\ref{metrics_all}). This indicates that the regions visited by both algorithms tend to be closer to each other. On the other hand, for the harder DASCMOP6, the trajectories do not overlap, and the number of shared nodes is much smaller, 14 for both cases. This indicates a high impact of the problem difficulty on the search ability of the MOEA/D family and that the MOEAs are visiting unrelated regions in the decision space. 

In terms of best Pareto optimal solutions and for the DASCMOP1, auto-MOEA/D was able to find one Pareto optimal solution while the no-restart variant found two and the update variant, none. Given the distribution of the nodes in the STNs, these optimal solutions might not be easy to find. No MOEA/D reached an optimal solution in the DASCMOP6. This supports our decision of separating the DASCMOP benchmark into two different groups, according to their difficulties.

Although we only showed the STNs of these two pairs, an identical trend occurs for all pairs of MOEAs: the STNs of all pairs of variants for the easy group are visually similar among themselves and the STNs, but different to the STNs of the hard group. Finally, looking at Table~\ref{metrics_all} we see the same results in terms of shared nodes; in the easy MOPs, there are more shared nodes, while for the hard MOPs, there are fewer shared nodes.

\section{Conclusion}
\label{discussion}

The aim of this work was to introduce a new methodology that investigates the effects of different configurations from the component-wise point of view of automatically designed multiobjective algorithms in terms of their decision space dynamics in constrained MOPs. We contrasted the behaviour of these configurations in terms of how they explore the decision space by comparing their Search Trajectory Networks (STNs), their population diversity, and, for reference, their hypervolume. This analysis allowed us to identify what the most influential components are.

We studied the auto-MOEA/D, a well-performing MOEA/D designed by the irace package, and the subsequently derived variants that differ from this machine-designed MOEA by a single component. The results have shown that the most influential components are the restart and update strategies. Their relative influence depends on the problem difficulty: not using the restart strategy was more influential in problems where MOEA/D performs well. In contrast, the update strategy was more influential in problems where MOEA/D performs poorly.

We also conduct a brief study on the effects of the decision space metrics. We found that most of the STN metrics can provide precise information about the search dynamics of the different MOEAs, with the only exception being the number of edges. Moreover, we found a high correlation between the IGD metric and: the number of nodes, edges and shared nodes; and the population variance. This result indicates that our decision space metrics can provide insightful information about the dynamics of different MOEAs in the decision space while also correlating to increments in performance.

Overall, this study strengthens the view that characterizing the effects of MOEA/D algorithm components could help in developing even more effective MOEAs. Taken together, these findings suggest a role for improving in promoting the study of specific components to develop new and better components, especially in consideration of the restart and update strategy.

We understand that our results are of interest to the broad multiobjective evolutionary computation community. 
One limitation of our work is that we only consider first-order interactions between the MOEA/D variants and their components. A natural progression includes considering higher-order interactions and conducting a similar analysis on other multiobjective metaheuristics, such as the NSGA-II solver.

\section*{Errata}

The authors found errors in the scripts for constructing and visualising the STNs.  We have updated the code repository, some tables, figures and corresponding text with the corrected STNs.  The general conclusions of the paper remain the same.

\clearpage

\bibliographystyle{ACM-Reference-Format}
\bibliography{references}

\end{document}